\documentclass[runningheads]{llncs}
\usepackage[T1]{fontenc}
\usepackage{amsmath}
\usepackage{amssymb}

\usepackage{graphicx}
\usepackage{float}
\usepackage{subcaption}
\usepackage{adjustbox}

\usepackage{booktabs}
\usepackage{multirow}
\usepackage{tablefootnote}

\usepackage{algorithm}
\usepackage{algorithmicx}
\usepackage{algcompatible}
\usepackage{algpseudocode}
\usepackage{url}
\usepackage{xcolor}
\usepackage{pifont}
\usepackage{hyperref}
\usepackage{comment}

\usepackage[misc]{ifsym}

\usepackage{mwe}
\usepackage{tikz}
\usetikzlibrary{arrows.meta}

\usepackage{xcolor}

\begin{document}

\title{UniFair: A unified fair clustering approach based on separation and compactness}

\titlerunning{UniFair: A unified fair clustering approach}
% If the full title of your paper is short enough to also fit in the running head, you can omit the abbreviated paper title here. You can check as follows: if you comment out the \titlerunning line, something will appear in the header of all odd-numbered pages of your PDF from page 3 onward. This something is either the full title (in which case all is well), or the error message "Title Suppressed Due to Excessive Length". If this error message appears, you're going to want to provide an abbreviated title within the \titlerunning command, because if you won't do it, Springer will do it for you.

%N.B.: Author information (both in the \author{} and \authorrunning{} command) should only be present in the Camera-Ready Version of your paper. The version that you initially submit for review, ought to be double-blind. So, when initially submitting your paper, use:
%\author{Author information scrubbed for double-blind reviewing}
\author{ Antonia Karra\inst{1} \and Vasiliki Papanikou\inst{1,2} \and Georgios Vardakas\inst{1} \and Evaggelia Pitoura\inst{1,2} \and Aristidis Likas\inst{1} }
% You may leave out the orcidID information, if you want to.
% Use \corr to indicate the corresponding author. Note the spacing around the \corr command. Only one author can be the corresponding author.

%N.B.: comment out the \authorrunning{} command for the double-blind version of your paper submitted for review. Later, if your paper is accepted, use the command for the Camera-Ready Version.
\authorrunning{Antonia Karra et al.}
% If there are two authors, write 'A.L. Benjamin and C.C. Broadus Jr.'
% If there are more than two authors, '[...] et al.' is used.

\institute{University of Ioannina, Greece \email{\{a.karra, v.papanikou, g.vardakas, pitoura, arly.cs\}@uoi.gr}
\and
Archimedes/Athena Research Center, Greece}

\maketitle              % typeset the header of the contribution

\begin{abstract}
Clustering is increasingly used to support high-impact decisions, yet standard objectives such as $k$-means can produce clusterings that treat demographic groups unequally. Existing fair clustering methods typically optimize a single notion of fairness and often overlook how clustering costs interact with the geometry of the induced decision boundaries. We propose \textsc{UniFair}, a unified framework that jointly optimizes \emph{separation fairness} and \emph{social fairness}. Separation fairness encourages protected groups to lie farther from the induced decision boundaries, while social fairness reduces disparities in within-cluster distortion by penalizing group-wise clustering costs. We develop gradient-based optimization procedures for separation-fair and unified $k$-means objectives, and extend them to deep clustering by enforcing the same criteria in the latent space of an autoencoder. Experiments on tabular and image datasets show that \textsc{UniFair} reduces both boundary-related and cost-based group disparities with only a modest increase in clustering loss.

\end{abstract} 
\section{Introduction}
\label{sec:intro}
Clustering is a core task in data analysis and machine learning, widely used for exploration, profiling, and as a building block for downstream decision support. In settings involving protected groups, standard clustering objectives can produce clusters that treat groups unequally, even when the overall clustering quality is high \cite{abbasi2021fair,ghadiri2021socially,chierichetti2017fair,li2020deep}. Since
the induced partition 
%becomes a structural assumption that 
shapes later processing and  decisions, group-level disparities in the resulting clusters can propagate to subsequent analyses and workflows.

Recent research on fair clustering proposes multiple fairness notions~\cite{overview}. One line of research aims to balance group proportions within clusters~\cite{ahmadi2020fair,chierichetti2017fair} and to provide guarantees on group presence across clusters~\cite{minirel}. Another line of research enforces a form of social fairness by controlling the clustering costs incurred by each group~\cite{abbasi2021fair,ghadiri2021socially}. However, disparities may also arise from how decision boundaries intersect groups. Two groups may incur similar clustering costs while one lies systematically closer to cluster boundaries, making its assignments more sensitive to small perturbations. This boundary-proximity effect captures a geometric source of disparity that is not reflected by distortion alone.

To address this gap, we introduce \textit{separation fairness}, a boundary-aware group fairness notion that encourages protected groups to lie farther away from their closest decision boundaries. Our formulation builds on a \textit{counterfactual-distance} view of boundary proximity: for each point, we measure the minimum perturbation required to reach the nearest separating hyperplane between its assigned centroid and its closest competing centroid. 
Separation fairness improves the least-separated group by maximizing its average distance to the decision boundary. This notion is complementary to cost-based fairness objectives: it targets the geometry of the partition rather than only the compactness of clusters.
We focus on $k$-means clustering, but our notion is applicable to other clustering objectives that induce well-defined decision boundaries between clusters.

Building on separation fairness, we propose \textsc{UniFair}.
%, which
%jointly optimizes boundary-based and cost-aware group fairness. 
\textsc{UniFair} couples a separation-fairness term with a social-fairness term, thereby capturing two complementary sources of disparity: proximity to the induced decision boundaries and disparities in within-cluster distortion.
We develop Lloyd-style alternating procedures, augmented with fairness-aware centroid updates, for (i) a \textit{separation fair $k$-means} objective and (ii) the unified \textsc{UniFair} objective.
%that jointly optimizes separation and social fairness. 
Our updates preserve the standard assignment--update structure of $k$-means while incorporating additional terms that (a) push groups farther from induced decision boundaries and (b) penalize group-wise clustering costs. To support high-dimensional data, we further extend the unified objective to a deep latent-space setting via an autoencoder, enabling joint optimization of reconstruction error, latent compactness, and fairness regularization terms.

Our empirical evaluation on tabular and image datasets demonstrates that optimizing separation fairness increases the minimum group separation from induced decision boundaries, while \textsc{UniFair} achieves strong joint improvements reducing both boundary-related and cost-based group disparities with only a modest increase in clustering loss. We further show that separation fairness and social fairness capture complementary aspects of unfairness, and that our algorithm provides a flexible way to balance the two fairness notions through the fairness weights, while remaining competitive with existing fair clustering and deep fair clustering methods.

The rest of this paper is structured as follows:
Section~\ref{sec:related} reviews related work.
Section~\ref{sec:separation_fairness} introduces separation fairness and the unified objective, while
Section~\ref{sec:algorithms} presents the separation and the \textsc{UniFair} algorithms.
Section~\ref{sec:experiments} presents our experimental evaluation, and Section~\ref{sec:conclusions} concludes the paper. 
\section{Related Work}
\label{sec:related}
Fairness in clustering has received considerable attention in recent years~\cite{overview}. 
At a high level, fairness notions in clustering can be divided into \textit{individual} and \textit{group} fairness. 
Individual fairness approaches, e.g.,~\cite{anderson2020distributional,kleindessner2020notion}, require that similar individuals receive similar cluster assignments, regardless of group membership. 
Group fairness approaches, in contrast, impose fairness requirements at the level of protected groups. 
Our work belongs to the latter category.

Several notions have been proposed to formalize group fairness in clustering, including \textit{balance-based}, \textit{representation-based}, and \textit{social-based} approaches. 
Balance-based fairness, e.g.,~\cite{chierichetti2017fair,ahmadi2020fair,schmidt2018fair,bera2019fair}, seeks to preserve the proportion of individuals from each protected group within clusters, thereby avoiding over- or under-representation. 
Minimum representation-based fairness, e.g.,~\cite{minirel}, requires that each group attain a sufficient level of representation in the clusters.

The line of work most closely related to ours is \textit{social fairness} in clustering. Rather than focusing on cluster composition, social fairness focuses on whether the quality of the clustering outcome is distributed fairly across groups. 
This is typically captured through group-aware clustering objectives based on the distances of points to their assigned centers, with the goal of preventing some groups from systematically receiving worse clustering losses than others.
Socially fair $k$-means~\cite{ghadiri2021socially} extends the standard $k$-means objective by optimizing a fairness-aware aggregation of group-specific costs, thus emphasizing the worst-off groups.
The method follows a Lloyd-style iterative refinement procedure with fairness-aware group-level objectives.
A related formulation is equitable group representation clustering~\cite{abbasi2021fair}, which evaluates fairness through how well the selected centers represent the data distribution of each group, introducing absolute and relative representation errors to quantify group-wise disparities. 
Subsequent work has further advanced this direction by developing improved approximation algorithms for group-fair clustering objectives, e.g.,~\cite{makarychev2021approximation,fpt}.
 
Our work captures a different aspect of group unfairness. 
Whereas social-fairness focus on group-wise clustering quality, we focus on group-wise \emph{separation}. 
Specifically, we ask whether points from some groups are systematically closer to cluster boundaries, and are therefore more likely to change assignment under small perturbations. 
Thus, our notion captures disparities in \emph{assignment stability} across groups, rather than disparities in group-wise clustering quality.

Finally, deep fair clustering incorporates fairness constraints into deep representation learning.
%to mitigate biases that traditional clustering methods may inherit from the data. 
Fairoids~\cite{wang2019towards} introduces fairness reference points defined as the average latent embeddings of the protected groups. It encourages cluster centroids to remain equidistant from these fairoids and reduces distributional disparities using Maximum Mean Discrepancy (MMD).
%, thereby promoting fairer group representation in the learned space. 
The work in~\cite{zhang2021deep} further develops this direction by combining probabilistic discriminative clustering with integer linear programming constraints to enforce group-level fairness. 
%It iteratively refines cluster assignments through pseudo-labeling and uses contrastive learning to improve the robustness of the learned representations. 
Similarly, the deep fair clustering approach (DFC) in~\cite{li2020deep} promotes statistical independence between cluster assignments and sensitive attributes through an adversarial learning framework that combines fairness-adversarial, structure-preserving, and clustering regularization losses. 

Overall, these deep fair clustering approaches primarily focus on equitable representation or independence with respect to sensitive attributes, rather than on 
social cost and separation fairness. %as in our work.
%group-wise clustering quality or assignment stability. In contrast, in this paper, we show how to incorporate social and separation fairness constraints into deep clustering.

\section{Separation and Unified Fairness}
\label{sec:separation_fairness}

Let $X = \{x_1, \dots, x_N\} \subset \mathbb{R}^d$ denote the dataset. We assume that the data points are partitioned into $m$ groups
$X_g$, $1 \leq g \leq m$.
Let \( M = \{\mu_1, \mu_2, \dots, \mu_k\} \) denote the set of \( k \) cluster centroids and let  $L(M)
=
\frac{1}{|X|}
\sum_{j=1}^k
\sum_{x \in C_j}
\|x - \mu_j\|^2$ be the clustering cost.

Previous research on fairness for clustering has mainly focused on the  clustering cost for each group. Specifically, for each group $X_g$, $1 \leq g \leq m$, the group-wise clustering cost is defined as:
\begin{equation}
\label{eq:social-fairness-term}
L_g(M)
=
\frac{1}{|X_g|}
\sum_{j=1}^k
\sum_{x \in C_j \cap X_g}
\|x - \mu_j\|^2.
\end{equation}

Social fairness introduced in \cite{ghadiri2021socially} focuses on the clustering cost incurred for each group. Specifically, social fairness focuses on the
 the maximum group-wise clustering cost incurred by any of the groups. 

\begin{definition}[Social Cost]
The social cost of clustering $M$ is the maximum group-wise clustering cost:
\end{definition}
\begin{equation}
Soc(M) = \max_{1 \leq g \leq m} L_g(M).
\label{eq:social_cost}
\end{equation}

%We would like to minimize this loss.
Social cost admits a complementary interpretation.
Clustering naturally induces a local partition of each group within each cluster.
For each cluster $C_j$ and group $X_g$, $1 \leq g \leq m$, we define its group mean as:
$
\mu_{j,g}
=
\frac{1}{|C_j \cap X_g|}
\sum_{x \in C_j \cap X_g} x.
$
The subgroup mean describes how the  group is represented locally inside cluster $C_j$.
The inner sum of Eq. \ref{eq:social-fairness-term} can be written as:
\begin{equation} 
\label{eq:social-fairness-equivalent} 
\sum_{x \in C_j \cap X_g} \|x - \mu_j\|^2 = \sum_{x \in C_j \cap X_g} \|x - \mu_{j,g}\|^2 + |C_j \cap X_g| \; \|\mu_j - \mu_{j,g}\|^2. \end{equation}

The second term shows that minimizing $L_g$ encourages each centroid $\mu_j$ to align
with the subgroup mean $\mu_{j,g}$, while the first term captures within-group dispersion.

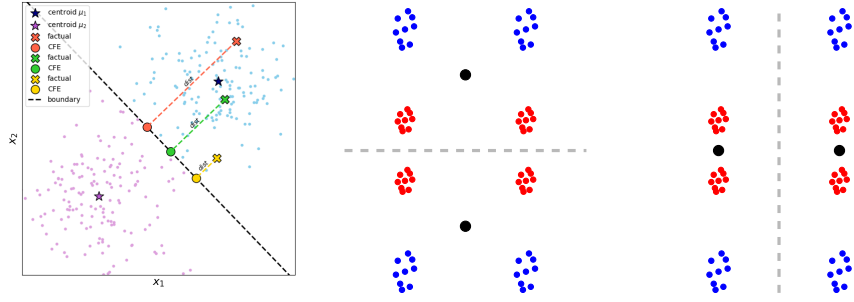
\begin{figure}[t]
\centering

\begin{subfigure}[t]{0.32\textwidth}
\centering
\includegraphics[width=\linewidth]{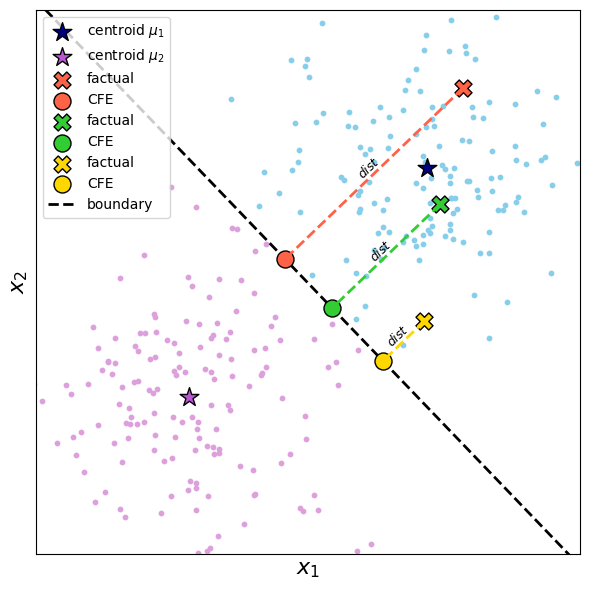}
\label{fig:k-means}
\end{subfigure}
\hfill
\begin{subfigure}[t]{0.32\textwidth}
\centering
\begin{tikzpicture}[scale=0.8]

\clip (-2.0,-2.35) rectangle (2.0,2.35);

% boundary y=0
\draw[dashed, very thick, gray!55] (-2.0,0) -- (2.0,0);

% centroids
\filldraw[black] (0,1.25) circle (2.4pt);
\filldraw[black] (0,-1.25) circle (2.4pt);

% Group A
\foreach \cx/\cy in {-1/2, 1/2, -1/-2, 1/-2}
{
  \foreach \dx/\dy in {
    0/0, 0.12/0.20, -0.12/0.18, 0.08/-0.25, -0.08/-0.20,
    0.15/0.05, -0.15/-0.05, 0.05/0.30, -0.05/-0.30}
  {
    \filldraw[blue] ({\cx+\dx},{\cy+\dy}) circle (1.25pt);
  }
}

% Group B
\foreach \cx/\cy in {-1/0.5, 1/0.5, -1/-0.5, 1/-0.5}
{
  \foreach \dx/\dy in {
    0/0, 0.08/0.12, -0.08/0.10, 0.06/-0.15, -0.06/-0.12,
    0.12/0.02, -0.12/-0.02, 0.04/0.18, -0.04/-0.18}
  {
    \filldraw[red] ({\cx+\dx},{\cy+\dy}) circle (1.25pt);
  }
}

\end{tikzpicture}
\end{subfigure}
\hfill
\begin{subfigure}[t]{0.32\textwidth}
\centering
\begin{tikzpicture}[scale=0.8]

\clip (-2.0,-2.35) rectangle (2.0,2.35);

% boundary x=0
\draw[dashed, very thick, gray!55] (0,-2.35) -- (0,2.35);

% centroids
\filldraw[black] (-1,0) circle (2.4pt);
\filldraw[black] (1,0) circle (2.4pt);

% Same dataset
\foreach \cx/\cy in {-1/2, 1/2, -1/-2, 1/-2}
{
  \foreach \dx/\dy in {
    0/0, 0.12/0.20, -0.12/0.18, 0.08/-0.25, -0.08/-0.20,
    0.15/0.05, -0.15/-0.05, 0.05/0.30, -0.05/-0.30}
  {
    \filldraw[blue] ({\cx+\dx},{\cy+\dy}) circle (1.25pt);
  }
}

\foreach \cx/\cy in {-1/0.5, 1/0.5, -1/-0.5, 1/-0.5}
{
  \foreach \dx/\dy in {
    0/0, 0.08/0.12, -0.08/0.10, 0.06/-0.15, -0.06/-0.12,
    0.12/0.02, -0.12/-0.02, 0.04/0.18, -0.04/-0.18}
  {
    \filldraw[red] ({\cx+\dx},{\cy+\dy}) circle (1.25pt);
  }
}

\end{tikzpicture}
\end{subfigure}

\caption{(a) Example of counterfactual distances. (b) Socially fair but separation unfair clustering. (c) Separation fair but socially unfair clustering.}
\label{fig:clustering_fairness}
\end{figure}

Social fairness controls within-cluster distortion across groups; however, it does not directly control how close each group lies to the induced decision boundaries. As shown in Figure~\ref{fig:clustering_fairness}(b), two groups may incur similar group-wise clustering costs, while one group lies  closer to the cluster boundary than the other. Points close to a boundary can change clusters under relatively small perturbations, whereas points farther away require larger changes.
In many applications, cluster assignments influence downstream decisions,
such as resource allocation, profiling, or recommendation.
If members of one group are systematically closer to decision boundaries,
their assignments become more sensitive to small perturbations or noise.
This can result in higher instability or volatility of the outcomes for that group.
We therefore introduce \textit{separation fairness} to explicitly quantify group-level separation from decision boundaries. Separation fairness complements distortion-based notions by capturing a geometric aspect of fairness that is not reflected by within-cluster distortion alone.

The centroids $M=\{\mu_1,\dots,\mu_k\}$ induce a Voronoi partition of
$\mathbb{R}^d$, where each cluster $C_j$ corresponds to the Voronoi cell
\[
C_j=\{x\in\mathbb{R}^d:\|x-\mu_j\|\le \|x-\mu_\ell\|,\ \forall \ell\neq j\}.
\]
Decision boundaries between clusters are therefore Voronoi bisectors
between pairs of centroids.

For a point $x$, let $\mu_{j_1(x)}$ and $\mu_{j_2(x)}$ denote its two
closest centroids.
The relevant decision boundary is the Voronoi bisector between these
two centroids, consisting of all points $z$ satisfying
$
\|z-\mu_{j_1(x)}\|^2=\|z-\mu_{j_2(x)}\|^2.
$
This bisector is a hyperplane passing through the midpoint
$
m(x)=\tfrac12(\mu_{j_1(x)}+\mu_{j_2(x)})
$
and orthogonal to
$
v(x)=\mu_{j_2(x)}-\mu_{j_1(x)}.
$

We define the counterfactual distance of $x$ as its squared Euclidean
distance to this Voronoi bisector:
$
\mathrm{dist}^2(x)
=
\big((x-m(x))^\top \hat v(x)\big)^2,$
where
$\hat v(x)=\frac{v(x)}{\|v(x)\|}.
$
This quantity corresponds to the squared Voronoi margin of $x$. Intuitively, this captures how stable a cluster assignment is for a point:
points close to the decision boundary can switch clusters under small perturbations,
whereas points farther away require larger changes.
We call this distance \emph{counterfactual} because it measures the smallest perturbation that would change the cluster assignment for the point (factual), aligning with counterfactual explanations (CFE) for clustering~\cite{VardakasKPL25}.
We use squared boundary distance for consistency with the squared
$k$-means objective and to obtain smooth gradients within the assignment regions. 

An example of counterfactual distances is  shown in Figure \ref{fig:clustering_fairness} (a).

\begin{comment}
\begin{figure}[ht]
    \centering
    \includegraphics[width=0.3\linewidth]{Figures/k-means_CFEs_no_constraints_2.png}
    \caption{Example of counterfactual distances.}
    \label{fig:k-means}
\end{figure}
\end{comment}

\iffalse
For each point \( x \), we denote its cluster and its nearest cluster as:
\[
j_1(x) = \arg\min_{j} \lVert x - \mu_j\rVert, 
\quad
j_2(x) = \arg\min_{j \neq j_1(x)} \lVert x - \mu_j\rVert,
\]
and the corresponding centroids as \( \mu_{j_1(x)} \) and \( \mu_{j_2(x)} \).

The separating hyperplane between centroids 
$\mu_{j_1(x)}$ and $\mu_{j_2(x)}$ 
consists of all points $z$ satisfying:
\begin{equation}
    |z - \mu_{j_1(x)}|^2 = |z-\mu_{j_2(x)}|^2
\end{equation}
This equation defines a \emph{hyperplane} that is perpendicular at the middle point:
\[m(x) = \tfrac{1}{2}\big(\mu_{j_1(x)} + \mu_{j_2(x)}\big)\] to the vector
\[v(x) = \mu_{j_2(x)} - \mu_{j_1(x)}\] 
connecting the two centers. We refer to the cluster boundary as \emph{separating hyperplane}.

We define the \textit{counterfactual distance} of a point $x$ as its squared distance to the perpendicular hyperplane equidistant from its two closest centroids:
\[
\mathrm{dist}^2(x) = \big((x - m(x))^\top \hat v(x)\big)^2,
\quad
\hat v(x) = \frac{v(x)} {\lVert v(x)\rVert}
\]
\fi

Let $d_1(x)=\|x-\mu_{j_1(x)}\|$ and
$d_2(x)=\|x-\mu_{j_2(x)}\|$ denote the distances to the closest and
second-closest centroids, respectively.
The counterfactual distance can then be written as:
$
\mathrm{dist}^2(x)=
\frac{\big(d_2^2(x)-d_1^2(x)\big)^2}
{4\|\mu_{j_2(x)}-\mu_{j_1(x)}\|^2},
$
which highlights its interpretation as a normalized geometric margin
between the two closest centroids.
The above expression is invariant to translations and rotations.
Under uniform scaling by a factor $s>0$, it scales as
$\mathrm{dist}^2(x)\mapsto s^2\,\mathrm{dist}^2(x)$,
consistent with the squared $k$-means objective.
Normalization by $\|\mu_{j_2(x)}-\mu_{j_1(x)}\|^2$
ensures that separation is measured relative to centroid spacing.

Note that although the counterfactual distance is expressed in terms of the Voronoi structure induced by $k$-means,
the notion of separation fairness applies to any nearest-centroid partition,
and can be extended to alternative clustering models that admit well-defined decision boundaries.

For each subgroup $X_g$, \( 1 \leq g \leq m \), we define the \emph{group counterfactual distance} as the average of the counterfactual distances of its group members:
\begin{equation}
\mathrm{cfd}_g(M) = \frac{1}{|X_g|}\sum_{x \in X_g} \big((x - m(x))^\top \hat{v}(x)\big)^2.
\label{eq:seperation-fairness-term}
\end{equation}
We use the average margin to capture typical assignment stability within each group,
avoiding sensitivity to isolated boundary points.

We use group counterfactual distances to define separation fairness. Our objective is to improve the boundary separation of the group that is currently least separated from the induced decision boundaries. 

\begin{definition}[Separation Fairness]
Separation fairness of clustering $M$ is:
\begin{equation}
Sep(M)=\min_{1 \leq g \leq m} \mathrm{cfd}_g(M).
\label{eq:separation-fairness}
\end{equation}
\end{definition}

Using the minimum yields a worst-group fairness criterion: maximizing $Sep(M)$ improves
the average boundary distance of the group that is currently most exposed to assignment
instability.
Figure \ref{fig:clustering_fairness}(c) shows a clustering that is separation fair but social unfair, illustrating the complementarity of the two fairness perspectives.

We now combine \emph{both} criteria: encouraging greater separation from induced decision boundaries (separation fairness) and reducing disparities in within-cluster distortion across groups (social fairness).

\begin{definition}[\textsc{UniFair} clustering objective]\label{def:unifair}
We minimize the following objective over centroids $M$ and assignments $c(\cdot)$:
\begin{equation}
\label{eq:unified-objective}
UniF(M)=\frac{1}{N}\sum_{i=1}^N\|x_i-\mu_{c_i}\|^2
+ \lambda_{\text{soc}}\,\operatorname{Soc}(M)
- \lambda_{\text{sep}}\,\operatorname{Sep}(M).
\end{equation}
\end{definition}

The first term preserves overall clustering quality by minimizing within-cluster distortion.
The second term penalizes the worst-group distortion, thereby reducing imbalance in clustering loss across groups,
while the third term promotes separation by increasing the minimum group margin.
The weights $\lambda_{\text{soc}}$ and $\lambda_{\text{sep}}$ control the trade-off between distortion and separation,
which are expressed on comparable squared-distance scales.
Separation fairness complements loss-based criteria by promoting greater separation from the induced decision boundaries, a property not captured by distortion alone.
By coupling separation with distortion minimization, the objective discourages trivial solutions that increase boundary separation only by excessively spreading centroids, since such configurations incur higher clustering losses.

\section{Algorithms}
\label{sec:algorithms}
In this section, we start by presenting our algorithm for separation fairness, and then extend the algorithm to satisfy the unified objective. Finally, we present a deep fair clustering approach.
\begin{algorithm}[H]
\caption{Separation Fair $k$-means Algorithm}
\label{alg:sepfair-kmeans}
\small
\begin{algorithmic} [1]
\Require Dataset $X$, groups $X_g$ for $1 \leq g \leq m$, number of clusters $k$, number of iterations $T$, learning rate $\eta$, fairness weight $\lambda_{\text{sep}}$ 
\State Initialize centroids $M = \{\mu_1, \mu_2, \dots, \mu_k\}$
\For{$t = 1$ to $T$}
  \State \textbf{Assignment step:} Assign each point to its closest centroid:
  \[
    c(x) = \arg\min_j \|x - \mu_j\|^2.
  \]
  \State Form clusters $C_j = \{x \in X : c(x)=j\}$.
  \State \textbf{Fairness evaluation:} Compute $\mathrm{cfd}_g(M)$, for each group $g$,  $1 \leq g \leq m$.
  \State Select the active group:
  \[
    G = \arg\min_{1 \leq g \leq m}  \mathrm{cfd}_g(M).
  \]
  \State \textbf{Gradient computation:}
  \State \hspace{1em} Compute $\nabla_{\mu_j} L_{\text{$k$-means}}$ from the standard $k$-means cost using Eq.~\eqref{eq:kmeans-gradient}.
  \State \hspace{1em} Compute $\nabla_{\mu_j} \mathrm{cfd}_G(M)$ using Eq.~\eqref{eq:fairness-gradient}.
  \State \textbf{Centroid update:}
  \[
    \mu_j \gets \mu_j - \eta \left(
             \nabla_{\mu_j} L_{\text{$k$-means}}
      - \lambda_{\text{sep}} \nabla_{\mu_j} \mathrm{cfd}_G(M)
    \right),
    \qquad \forall j \in \{1,\dots,k\}.
  \]
\EndFor
\State \Return Final centroids $M$ and assignments $c(\cdot)$.
\end{algorithmic}
\end{algorithm}
\subsection[Separation Fair $k$-means Algorithm]{Separation Fair $k$-means Algorithm}

We propose a separation fair $k$-means algorithm that jointly optimizes clustering quality and separation fairness.
Specifically, we optimize the clustering-quality term together with the separation-fairness term $\operatorname{Sep}(M)$ of Definition~\ref{def:unifair}:
\begin{equation}
\label{eq:sep-fairness-objective}
SepF(M)=\frac{1}{N}\sum_{i=1}^N\|x_i-\mu_{c_i}\|^2
- \lambda_{\text{sep}}\,\operatorname{Sep}(M).
\end{equation}
Algorithm~\ref{alg:sepfair-kmeans} proceeds iteratively by first recomputing the cluster assignments and then updating the centroids using the combined gradient of the $k$-means cost and the separation-fairness term for the currently worst-off group. Since assignments and the identities of the two closest centroids can change across iterations, the overall objective is piecewise smooth; we therefore view the procedure as an alternating heuristic in the spirit of the Lloyd algorithm, augmented with fairness-aware centroid updates.
%Algorithm~\ref{alg:sepfair-kmeans} proceeds iteratively by first recomputing the cluster assignments and then updating the centroids using the combined gradient of the $k$-means loss and the separation-fairness term for the currently worst-off group.
%%%The algorithm (presented in detail in Algorithm\ref{alg:sepfair-kmeans}) minimizes the objective with respect to the cluster centers using gradient descent.  It alternates
%%between standard cluster assignments and fairness-aware centroid updates that
%account for the worst-off group at each iteration.
We next derive the gradients used in the centroid update step.

%This section first presents the optimization procedure for Fair $k$-Means using the hyperplane counterfactual distance, followed by the explicit gradient formulas required for updating the centroids.

\subsubsection*{Gradient of the $k$-means loss term.}
\label{sec:gradients}
For each  centroid \( \mu_j \), the gradient of the $k$-means loss term is:
\begin{equation}
\label{eq:kmeans-gradient}
\nabla_{\mu_j} L_{\text{$k$-means}}
=
\frac{2}{N}
\sum_{x \in C_j}
(\mu_j - x).
\end{equation}
%For centroid \( \mu_j \), the gradient of the $k$-means loss term is:
%\begin{equation}
%\label{eq:kmeans-gradient}
%\nabla_{\mu_j} L_{\text{kmeans}}
%= 2\,|C_j|\!\left(\mu_j - \frac{1}{|C_j|}\sum_{x\in C_j} x \right),
%\end{equation}
%where \( C_j \) is the set of points assigned to cluster \( j \).
\subsubsection*{Gradient of the separation fairness term.}
For each group $X_g$, $1 \leq g \leq m$, we define:
$
\mathrm{cfd}_g(M) = \frac{1}{|X_g|} \sum_{x \in X_g} r(x;M),$
where
$r(x;M) = \big((x - m(x))^\top \hat v(x)\big)^2,
$
, $m(x)$ is the midpoint between the two closest centroids to $x$, and $\hat v(x)$ is the unit vector pointing from the closest centroid to the second closest centroid.
For a single point \( x \), we define:
$
s(x) = \hat v(x)^\top(x - m(x)),$ 
where
$P(x) = I - \hat v(x)\hat v(x)^\top.
$

Let $j_1(x)$ and $j_2(x)$ denote the indices of the two centroids closest to $x$. For a fixed nearest-centroid configuration, the gradients of $r(x;M)$ with respect to these centroids are:

\begin{adjustbox}{max width=\linewidth}
\begin{minipage}{\linewidth}
\begin{equation}
%\label{eq:fairness-gradient}
\begin{aligned}
\nabla_{\mu_{j_1(x)}} r(x)
&= 2\,s(x)\left(\,-\frac{1}{\|v(x)\|}\,P(x)\,(x-m(x))\;-\;\tfrac12\,\hat v(x)\right),\\[4pt]
\nabla_{\mu_{j_2(x)}} r(x)
&= 2\,s(x)\left(\,\frac{1}{\|v(x)\|}\,P(x)\,(x-m(x))\;-\;\tfrac12\,\hat v(x)\right).
\end{aligned}
\end{equation}
\end{minipage}
\end{adjustbox}

While for all remaining centroids,
$
\nabla_{\mu_j} r(x) = 0
\qquad \text{for } j \notin \{j_1(x), j_2(x)\}.
$

Averaging over the points in group \(X_g\) yields:
\begin{equation}
\label{eq:fairness-gradient}
\nabla_{\mu_j}\,\mathrm{cfd}_g(M)
=
\frac{1}{|X_g|}
\sum_{x \in X_g}
\Big(
\mathbf{1}\{j = j_1(x)\}
+
\mathbf{1}\{j = j_2(x)\}
\Big)\,
\nabla_{\mu_j} r(x).
\end{equation}

Therefore, at each iteration, the fairness-aware update increases the average counterfactual distance of the currently worst-off group (under the current assignment and nearest-centroid configuration), while the $k$-means term promotes cluster compactness. Compared to vanilla $k$-means, the additional computation amounts to identifying the two closest centroids for each point and accumulating the corresponding group-wise gradients.

%\subsection{Objective Function}

%\subsection[Socially Fair k-Means Algorithm]{Socially Fair $k$-Means Algorithm}
\subsection{\textsc{UniFair} $k$-means Algorithm}
We now extend our algorithm to incorporate the social fairness $Soc(M)$ term. First, we show how a similar active-group approach can be used to incorporate the social fairness term, yielding the following objective:
\begin{equation}
\label{eq:social-objective}
SocF(M)=\frac{1}{N}\sum_{i=1}^N\|x_i-\mu_{c_i}\|^2
+ \lambda_{\text{soc}}\,\operatorname{Soc}(M).
\end{equation}

First, we derive the gradient of the social fairness term.

\subsubsection*{Gradient of the social fairness term.}

Let the worst-off group be:
\[
G = \arg\max_{1 \leq g \leq m} L_g(M).
\]

Then the active fairness gradient is
\begin{equation}
\label{eq:social-gradient}
\nabla_{\mu_j} L_g(M)
=
\frac{2}{|X_g|}
\sum_{x \in C_j \cap X_g}
(\mu_j - x)
=
\frac{2|C_j \cap X_g|}{|X_g|}
(\mu_j - \mu_{j,g}),
\end{equation}
which shows that the fairness penalty pulls $\mu_j$ toward the subgroup mean $\mu_{j,G}$ of the group currently incurring the highest clustering cost.

The Social Fair $k$-means algorithm works similarly to Algorithm \ref{alg:sepfair-kmeans}. The only differences are: (1) in the fairness evaluation step (Lines 5-6), we select the worst group $G$ based on social cost, (2) in line (9), we compute $\nabla_{\mu_j} L_g(M)$ using Eq.~\eqref{eq:social-gradient}, and (3) the centroid update (Line 10) uses:
\[
  \mu_j
  \gets
  \mu_j
  - \eta\left(
      \nabla_{\mu_j} L_{\text{$k$-means}}
     + \lambda_{\text{soc}} \cdot \nabla_{\mu_j} L_G(M)
    \right).
  \]

We now present the \textsc{UniFair} $k$-means algorithm (Algorithm~\ref{alg:unified}) that minimizes the unified objective.

\begin{algorithm}[H]
\caption{\textsc{UniFair} $k$-means Algorithm}
\label{alg:unified}
\small
\begin{algorithmic}[1]
 \Require Dataset $X$, groups $X_g$ for $1 \leq g \leq m$, number of clusters $k$, number of iterations $T$, learning rate $\eta$, fairness weights $\lambda_{\text{soc}}$, $\lambda_{\text{sep}}$
\State Initialize centroids $M = \{\mu_1, \mu_2, \dots, \mu_k\}$
\For{$i=1$ to $T$}
\State Assign $c(x)=\arg\min_j\|x-\mu_j\|^2$ and form clusters $C_j$
\State Compute $L_g(M)$ and $\mathrm{cfd}_g(M)$ for each group $1 \leq g \leq m$
\State Active groups:  $H = \arg\max_{1 \leq g \leq m}  L_g(M)$,  $G = \arg\min_{1 \leq g \leq m}  \mathrm{cfd}_g(M)$
%\State Compute $\nabla_{\mu_j} L_{\text{kmeans}}$, $\nabla_{\mu_j} L_H(M)$ $\nabla_{\mu_j} \mathrm{cfd}_G(M)$ 
\State Update centroids
\[
\mu_j\leftarrow\mu_j-\eta\Big(  \nabla_{\mu_j} L_{\text{$k$-means}} +\lambda_{\text{soc}} \nabla_{\mu_j} L_H(M) -\lambda_{\text{sep}} \nabla_{\mu_j} \mathrm{cfd}_G(M) \Big),\quad \forall j
\]
\EndFor
\State \Return $M$ and assignments $c(\cdot)$
\end{algorithmic}
\end{algorithm}

Increasing $\lambda_{\text{sep}}$ emphasizes separation fairness by increasing $\min_{g}\mathrm{cfd}_g(M)$, while increasing $\lambda_{\text{soc}}$ emphasizes social fairness by reducing the worst group-wise loss $\max_{g} L_g(M)$, typically at some cost in overall distortion. When both terms are active, the method jointly trades off cluster compactness, group-wise distortion disparities, and group-level separation from induced decision boundaries.

\subsection{Deep Fair Clustering}
\label{sec:deep_fair_clustering}
%
%To further enhance the representation quality and fairness of the clustering process, 
We extend the fairness-aware objective into a deep learning framework to support high dimensional non tabular data. 
The key idea is to integrate fairness regularization terms into the latent space 
learned by an \textit{autoencoder (AE)}, enabling the joint optimization of 
reconstruction, clustering compactness, and fairness. 
The encoder $E_\phi$ maps the input $x_i$ into a latent embedding $z_i = E_\phi(x_i)$, 
where fairness constraints are applied, while the decoder $D_\psi$ reconstructs the input 
to preserve information fidelity. 
The total loss takes the general form
\begin{equation}
\label{eq:deep_total_general}
\mathcal{L}_{\text{total}} =
\alpha \, \mathcal{L}_{\text{rec}} +
\beta \, \mathcal{L}_{\text{cmp}} +
\lambda \, \mathcal{L}_{\text{fair}},
\end{equation}
where $\mathcal{L}_{\text{rec}}$ is the autoencoder reconstruction loss that ensures reconstruction quality, 
$\mathcal{L}_{\text{cmp}}$ is the latent space $k$-means that enforces clustering compactness in latent space, 
and $\mathcal{L}_{\text{fair}}$ represents a fairness-specific regularization term. The coefficients $\alpha$ and $\beta$ control the relative importance of the reconstruction and clustering objectives, respectively, while $\lambda$ determines the strength of the fairness regularization.  
Depending on the fairness criterion, $\mathcal{L}_{\text{fair}}$ takes a different form, 
leading to three variants described below.

\paragraph{Deep Separation Fair Clustering.}
We enforce separation fairness directly in latent space. For each group $g$, let $Z_g=\{z_i: x_i\in X_g\}$ denote the embeddings of points in group $g$. For each $z_i$, let $\mu_{a_i}$ and $\mu_{b_i}$ denote its closest and second-closest latent centroids, respectively, and let $r(z_i;M)$ be the (squared) distance of $z_i$ to the induced decision boundary, defined as in Section~\ref{sec:separation_fairness}:
\[
r(z_i;M)=\frac{\big(\|z_i-\mu_{b_i}\|^2-\|z_i-\mu_{a_i}\|^2\big)^2}{4\|\mu_{a_i}-\mu_{b_i}\|^2}.
\]
The group counterfactual distance in latent space is then
\[
\mathrm{cfd}_g(Z)=\frac{1}{|Z_g|}\sum_{z_i\in Z_g} r(z_i;M),
\qquad
\operatorname{Sep}(Z)=\min_{1\le g\le m}\mathrm{cfd}_g(Z).
\]
The deep separation-fair objective rewards larger worst-group separation from the induced decision boundaries:
\begin{equation}
\label{eq:deep_sep_objective}
\mathcal{L}_{\text{sep}}
=
\alpha\,\mathcal{L}_{\text{rec}}
+
\beta\,\mathcal{L}_{\text{cmp}}
-
\lambda_{\text{sep}}\,\operatorname{Sep}(Z).
\end{equation}

\paragraph{Deep \textsc{UniFair} Clustering.}
We now incorporate the social fairness term in the latent space. For each group $g$, define the group-wise latent clustering loss
\[
L_g(Z)=\frac{1}{|Z_g|}\sum_{z_i\in Z_g}\|z_i-\mu_{c(z_i)}\|^2,
\qquad
\operatorname{Soc}(Z)=\max_{1\le g\le m} L_g(Z).
\]
Combining social and separation fairness yields the deep \textsc{UniFair} objective:
\begin{equation}
\label{eq:deep_unifair_objective}
\mathcal{L}_{\textsc{UniFair}}
=
\alpha\,\mathcal{L}_{\text{rec}}
+
\beta\,\mathcal{L}_{\text{cmp}}
+
\lambda_{\text{soc}}\,\operatorname{Soc}(Z)
-
\lambda_{\text{sep}}\,\operatorname{Sep}(Z).
\end{equation}

This objective is the latent-space analogue of the classical \textsc{UniFair} formulation, simultaneously penalizing worst-group distortion and rewarding worst-group separation from induced decision boundaries.
%Because assignments and the identities of the closest/second-closest centroids can change during training, the objective is only piecewise smooth. In practice, we alternate between (i) recomputing assignments and the corresponding closest/second-closest centroids in latent space and (ii) taking mini-batch gradient steps on $\phi$, $\psi$, and $M$ with these quantities held fixed. 
The weights $\lambda_{\text{soc}}$ and $\lambda_{\text{sep}}$ control the trade-off between the two criteria.

\section{Experimental Evaluation}
\label{sec:experiments}

For our experiments, we use four tabular datasets commonly used in fairness evaluation for clustering: Adult, Student, Bank, and Credit\footnote{\href{https://archive.ics.uci.edu/ml/datasets/adult}{Adult}, \href{https://archive.ics.uci.edu/dataset/320/student+performance}{Student}, \href{https://archive.ics.uci.edu/ml/datasets/Bank+Marketing}{Bank}, and \href{https://archive.ics.uci.edu/dataset/350/default+of+credit+card+clients}{Credit}}. Adult contains demographic and income information, Credit includes financial and behavioral attributes of credit card clients, Bank consists of marketing and demographic data from bank campaigns, and Student contains academic and personal background information. For Adult and Student we use \textit{sex} as the protected attribute, while for Bank and Credit we use \textit{marital status}. For deep fair clustering, we use the MNIST–USPS digit datasets\footnote{\href{http://yann.lecun.com/exdb/mnist/}{MNIST}
, \href{https://www.kaggle.com/datasets/bistaumanga/usps-dataset}{USPS}}
, as well as the Color Reverse MNIST dataset, where image colors are inverted. In MNIST–USPS, the dataset origin (MNIST or USPS) serves as the protected attribute, while in Color Reverse MNIST the image color (original or reversed) is treated as the protected attribute.

For tabular datasets, we use $k=7$ clusters for Adult and $k=5$ for the others. The number of clusters is selected based on the silhouette score. Optimization runs for $500$ iterations with learning rate $0.01$, and experiments are repeated over $10$ random seeds.% for $\lambda \in \{0, 0.2, 0.4, 0.6, 0.8, 1.0\}$.
For deep clustering, we use $k=10$ clusters and a fully connected autoencoder with latent dimension $20$ and LeakyReLU activations. The model is pretrained for $50$ epochs (learning rate $10^{-3}$) and then trained for clustering for $50$ epochs (learning rate $5\cdot10^{-5}$). The reconstruction and compactness weights are set to $\alpha=\beta=1.0$.
%For each \(\lambda\) we report (mean \(\pm\) st.d. across seeds) the $k$-means %cost, separation fairness, and social cost, as well as the separation fairness %gap and the social cost gap.
Our code is available online\footnote{\href{https://github.com/VasilikiPapanikou/UniFair-A-unified-fair-clustering-approach-based-on-separation-and-compactness.git}{GitHub code}}.

Through our experiments we address the following research questions:
\begin{itemize}
\item \textbf{RQ1:} What is the relationship between separation and social fairness?
\item \textbf{RQ2:} Does jointly optimizing separation and social fairness lead to
better overall equity?
\item \textbf{RQ3:} How does UniFair perform on multidimensional data?

\end{itemize}

%Through these research questions, we also study the effect of the fairness weight on clustering compactness and fairness objectives. For \textsc{UniFair}, we also examine how distributing the weight between separation fairness and social cost influences the resulting clustering. Finally, we analyze the loss evolution across epochs for different $\lambda$ values and compare our approach with existing fair clustering and deep fair clustering methods.

%\vspace*{0.1in}
%\noindent
\textbf{RQ1: What is the relationship between separation and social fairness?}

\begin{figure}[ht]
    \centering
    \includegraphics[width=\linewidth]{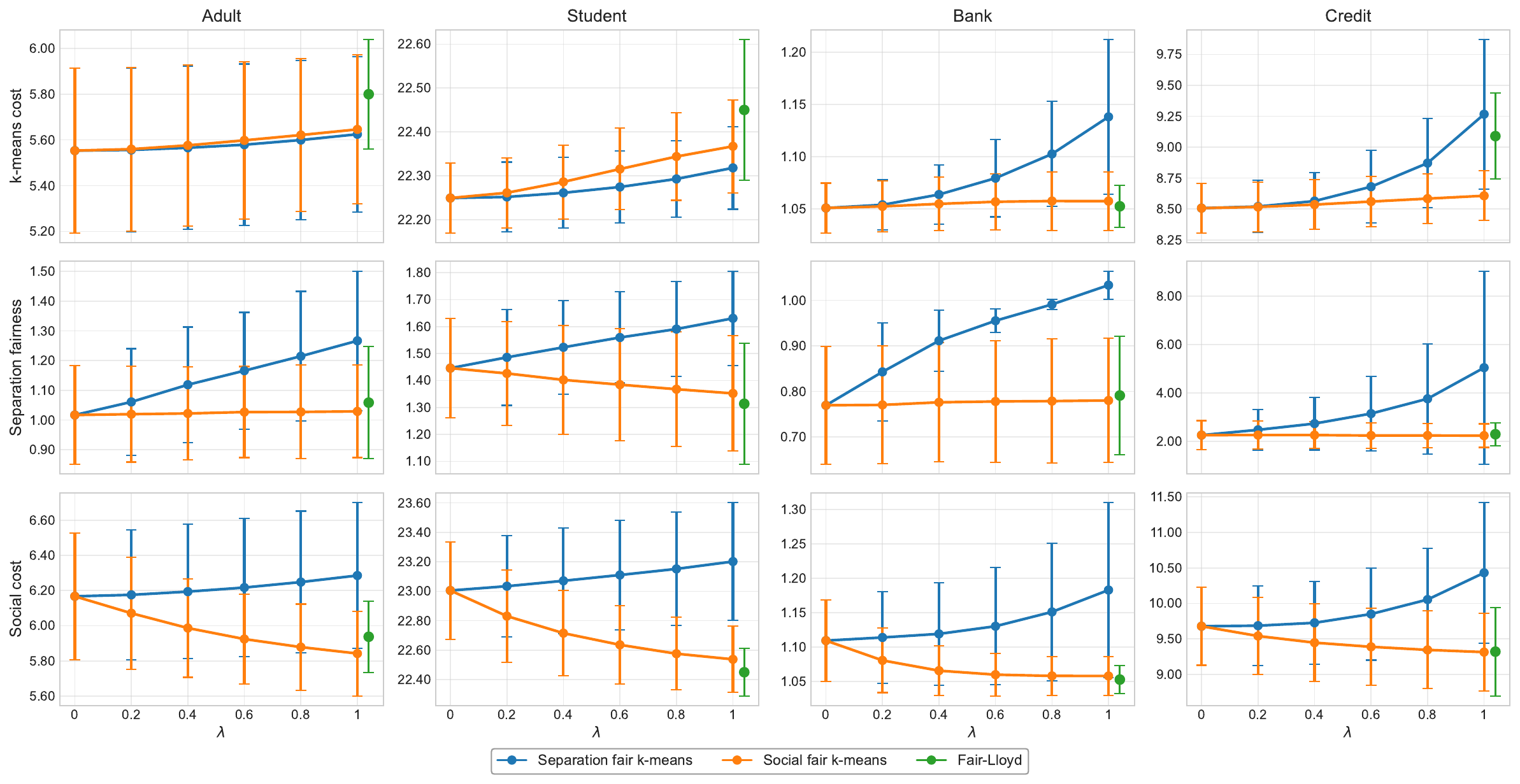}
    \caption{$k$-means cost and fairness costs for separation fair $k$-means, social fair $k$-means and Fair-Lloyd across datasets and $\lambda$ values.}
    \label{fig:cost_separation_social}
\end{figure}

\noindent In this research question, we investigate the relationship between separation fairness and social fairness. To this end, we evaluate both separation fair $k$-means and social fair $k$-means for different values of the fairness parameter $\lambda$, where $\lambda = \lambda_{sep} = \lambda_{soc}$ and we record the $k$-means clustering cost, the separation fairness, and the social fairness cost. The results for all datasets are presented in Figure~\ref{fig:cost_separation_social}. This experiment allows us to analyze two aspects. First, it reveals the trade-off between clustering quality and fairness as the fairness parameter increases. Second, it allows us to examine the interaction between the two fairness notions by observing how social fairness behaves when optimizing separation fairness, and conversely how separation fairness changes when optimizing social fairness.

\begin{figure}[ht]
    \centering
    \includegraphics[width=\linewidth]{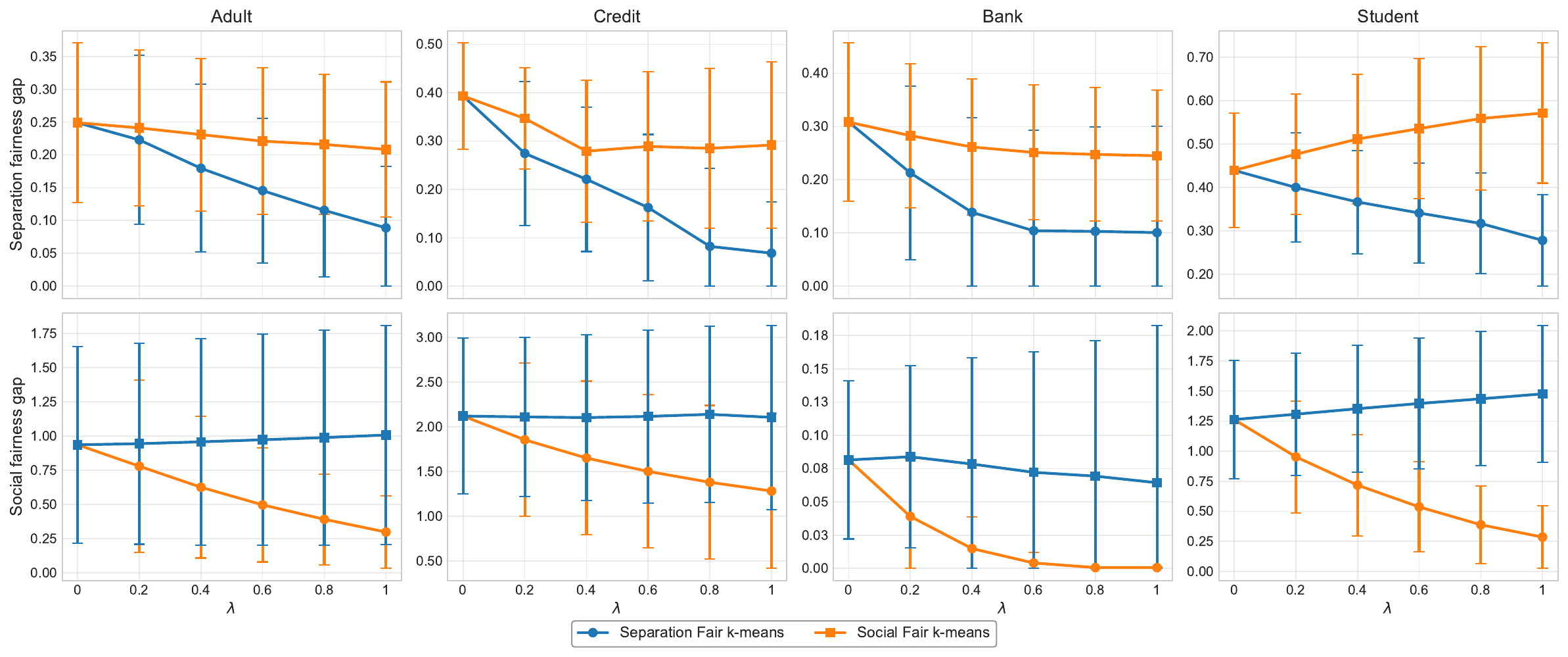}
    \caption{Separation and social fairness gaps in case of separation fair $k$-means and social fair $k$-means across datasets and $\lambda$ values .}
    \label{fig:fairness_gap_comparison}
\end{figure}

The results show that increasing the fairness parameter $\lambda$ has a limited impact on clustering performance as the 
$k$-means cost increases slightly when optimizing separation fairness, while for social fair $k$-means it remains nearly constant in most datasets.
Also, we see that the two fairness notions capture different types of unfairness.
Optimizing separation fairness improves the separation fairness 
as $\lambda$ increases, moving points from disadvantaged groups further from cluster decision boundaries, but may slightly increase the social cost.
Conversely, optimizing social fairness reduces the maximum social cost, leading to a more balanced clustering cost across demographic groups, while sometimes slightly worsening separation fairness.

In addition, in Figure~\ref{fig:fairness_gap_comparison} we report the separation fairness gap and social fairness gap for both algorithms. 
Although these quantities are not directly optimized by the objectives, they provide additional insight into disparities between demographic groups. We observe that the separation fair 
$k$-means algorithm steadily reduces the separation fairness gap as 
$\lambda$ increases, indicating that the distances of the two demographic groups from the cluster decision boundaries become more balanced. In contrast, the social fairness gap remains relatively stable and in some cases increases slightly. Similarly, when optimizing social fair 
$k$-means, the social fairness gap decreases across most datasets, while the separation fairness gap remains largely stable.
Overall, these results suggest that separation fairness and social fairness capture complementary aspects of fairness, and optimizing one notion does not significantly harm the other.

We also compare our approach with FairLloyd \cite{ghadiri2021socially}, which focuses on social fairness. While our method optimizes the fairness objective through gradient-based updates, FairLloyd enforces social fairness using a constrained centroid update that searches along the segment between the group means within each cluster. The results show that our social fair $k$-means algorithm achieves a lower $k$-means cost while maintaining comparable levels of social fairness. In some datasets, such as Adult, our method even achieves improved social cost. Moreover, the separation fairness achieved by FairLloyd is relatively low and comparable to that obtained by our social fair $k$-means algorithm.

\begin{figure}[ht]
    \centering
    \includegraphics[width=\linewidth]{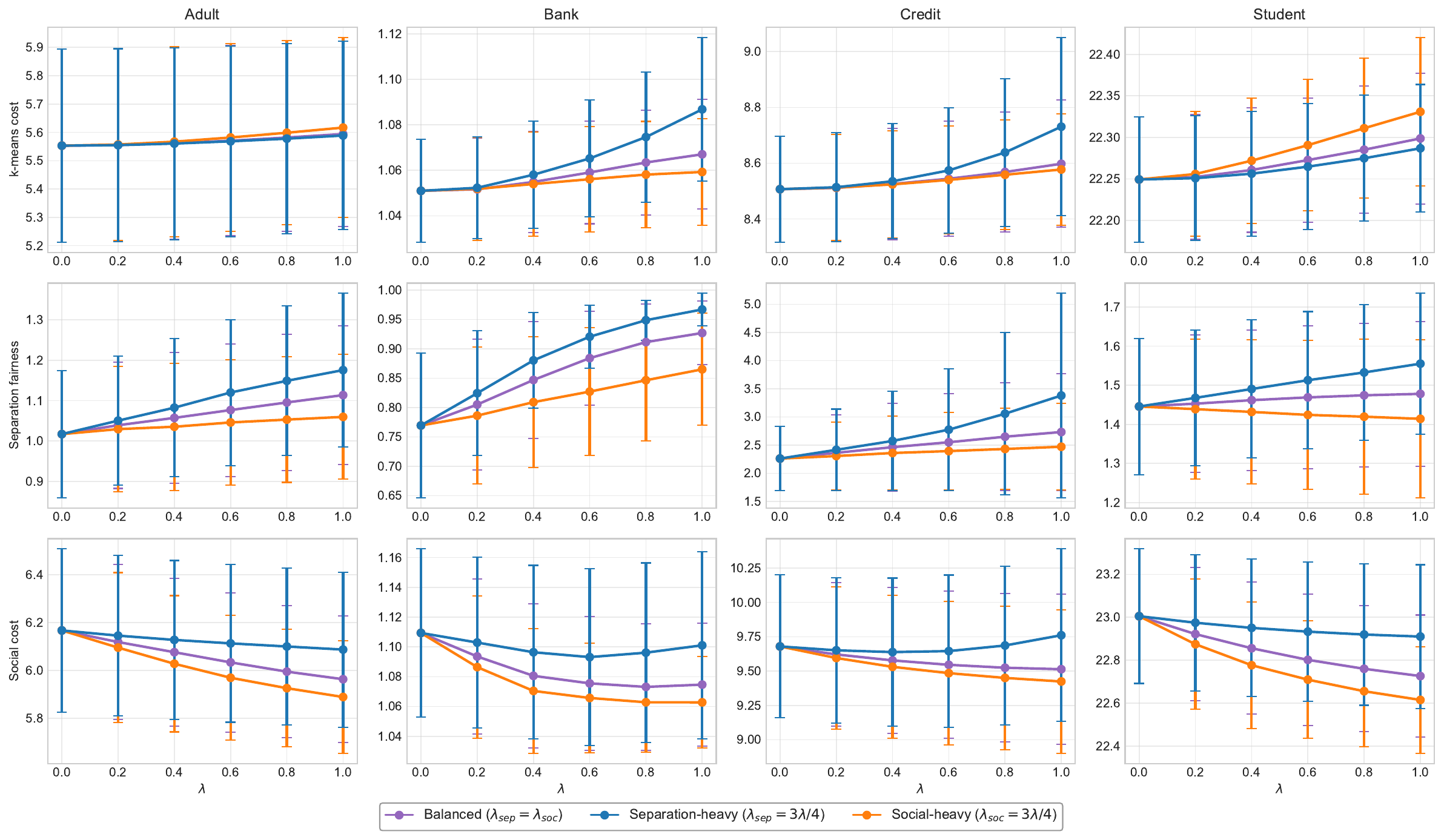}
    \caption{$k$-means cost, separation fairness and social cost for \textsc{UniFair} $k$-means across datasets and $\lambda$ values for different fairness weight configurations.}
    \label{fig:unifair_cost}
\end{figure}

%\noindent
\textbf{RQ2: Does jointly optimizing separation and social fairness lead to better overall equity?}

\noindent To examine this question, we evaluate our unified objective algorithm \textsc{UniFair}, which simultaneously optimizes separation and social fairness.
This allows us to study how different distributions of the fairness weight influence the trade-off between separation fairness and social fairness.
In this experiment, we vary the relative importance of the two fairness criteria and evaluate different settings of $\lambda_{sep}$ and $\lambda_{soc}$.
We consider three settings: a balanced case where $\lambda_{sep} = \lambda_{soc} = \lambda/2$, a setting with greater emphasis on separation fairness ($\lambda_{sep} = 3\lambda/4$), and a setting with greater emphasis on social fairness ($\lambda_{soc} = 3\lambda/4$).

Figure~\ref{fig:unifair_cost} shows the clustering cost, the separation fairness, and the maximum social cost, obtained by the \textsc{UniFair} algorithm across datasets and different values of $\lambda$ for the three weighting cases. 
\begin{comment}
\begin{figure}[ht]
    \centering
    \includegraphics[width=\linewidth]{Figures/unified_gaps_across_datasets.pdf}
    \caption{Separation and social gap for \textsc{UniFair} $k$-means across datasets and $\lambda$ values for different fairness weight configurations.}
    \label{fig:unifair_gaps}
\end{figure}
\end{comment}
Overall, we observe that the balanced setting ($\lambda_{sep}=\lambda_{soc}=\lambda/2$) achieves a good compromise between the two fairness notions, maintaining moderate values for both separation and social fairness while keeping the clustering cost relatively stable. 
When more weight is assigned to separation fairness ($\lambda_{sep}=3\lambda/4$), the separation fairness cost improves consistently as $\lambda$ increases, while the social fairness cost slightly increases. Conversely, when more weight is assigned to social fairness ($\lambda_{soc}=3\lambda/4$), the social fairness cost decreases, indicating improved balance between demographic groups, while the separation fairness cost tends to increase slightly.
\begin{figure}[ht]
    \centering
    \includegraphics[width=\linewidth]{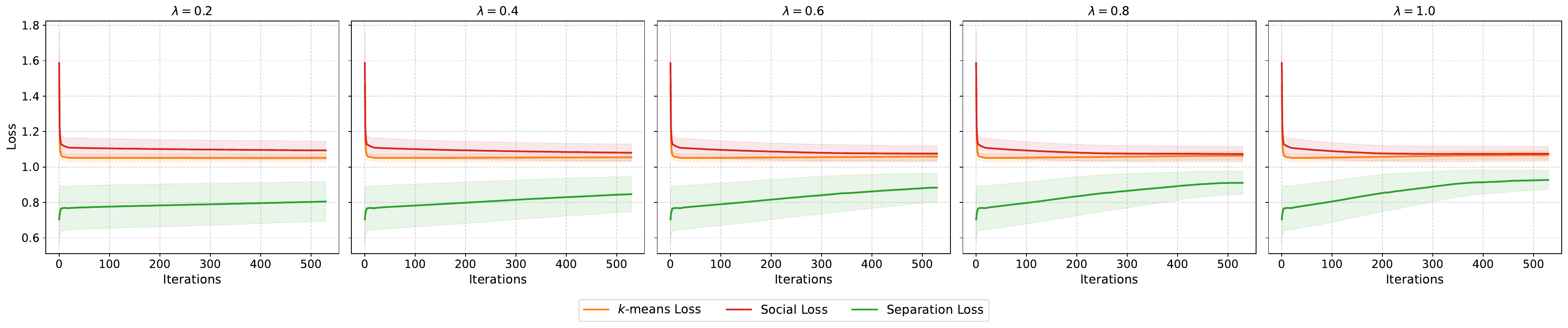}
    \caption{\textsc{UniFair} loss curves across iterations for $\lambda$ values on the Bank dataset.}
    \label{fig:loss_curves}
\end{figure}
%Figure~\ref{fig:unifair_gaps} reports the separation fairness gap and the social fairness gap achieved by \textsc{UniFair} across datasets and values of $\lambda$ for the different weighting schemes. When more weight is assigned to separation fairness ($\lambda_{sep}=3\lambda/4$), the separation gap decreases more significantly, while improvements in the social gap are more limited. Conversely, when more weight is assigned to social fairness ($\lambda_{soc}=3\lambda/4$), the social gap decreases more strongly, while the separation gap remains relatively stable. The balanced setting ($\lambda_{sep}=\lambda_{soc}=\lambda/2$) provides a compromise, where both fairness gaps decrease simultaneously.

Figure~\ref{fig:loss_curves} shows the evolution of the clustering, separation, and social losses of the \textsc{UniFair} objective during training for different values of $\lambda$ on the Bank dataset. 
We observe that the optimization process converges rapidly within the first iterations and stabilizes afterwards for all values of $\lambda$.
As $\lambda$ increases, the separation loss increases as it is maximized during optimization, while the social loss decreases. 
The clustering loss remains relatively stable across iterations and across different $\lambda$ values, indicating that improvements in fairness can be achieved without significantly affecting clustering quality.
Overall, these results demonstrate that jointly optimizing separation and social fairness improves fairness across both criteria while preserving clustering performance.

%\vspace*{0.1in}
%\noindent
\textbf{RQ3: How does \textsc{UniFair} perform on multidimensional  data?}

\begin{figure}[ht]
    \centering
    \includegraphics[width=\linewidth]{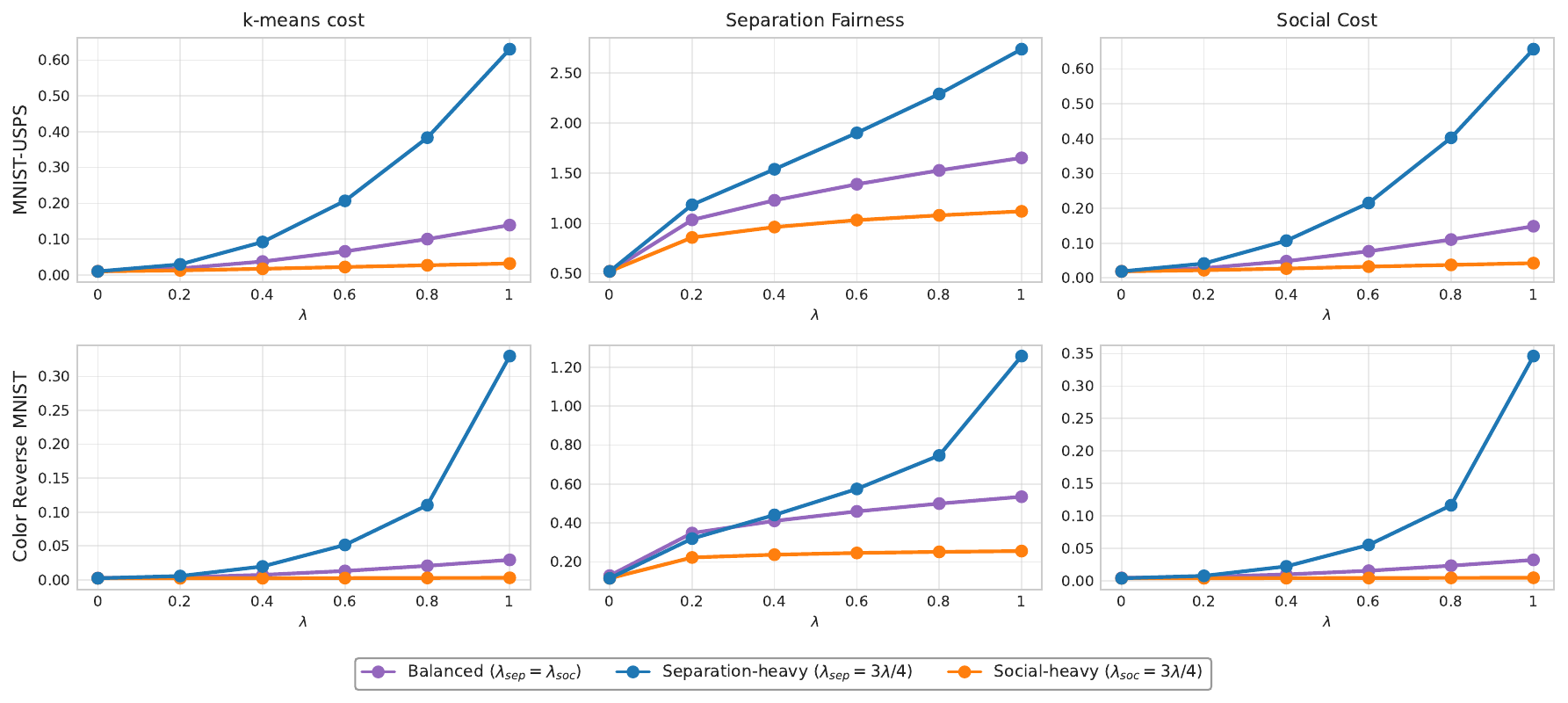}
    \caption{$k$-means cost, separation fairness and social cost for \textsc{UniFair} for different fairness weight configurations on MNIST-USPS and Color Reverse MNIST.}
    \label{fig:unifair_mnist_usps}
\end{figure}

\noindent To evaluate the effectiveness of \textsc{UniFair} on high-dimensional visual data, we conduct experiments on the MNIST–USPS and Color Reverse MNIST datasets. Figure~\ref{fig:unifair_mnist_usps} shows the $k$-means cost, separation fairness, and social cost across different values of $\lambda$ and for different fairness weight configurations.  

As $\lambda$ increases, fairness improves while the clustering cost rises moderately, particularly when separation fairness is emphasized. For both datasets, the social cost is already close to zero for small $\lambda$, indicating balanced clustering costs across groups even without strongly enforcing social fairness. Introducing separation fairness shifts cluster boundaries and centroids, which may slightly increase the social cost in the balanced setting. Assigning more weight to social fairness corrects these imbalances.
Overall, the fairness weights provide flexibility to adjust the relative importance of the two criteria depending on the dataset, allowing improvements in both fairness notions.
\begin{table}[t]
\centering
\caption{Comparison with the DFC method.}
\label{tab:deep_clustering_comparison}
\begin{tabular}{llccccc}
\toprule
& & & \multicolumn{2}{c}{Separation Fairness} & \multicolumn{2}{c}{Social Cost} \\
\cmidrule(lr){4-5} \cmidrule(lr){6-7}
Dataset & Method & NMI & Group A & Group B & Group A & Group B \\
\midrule

\multirow{2}{*}{MNIST-USPS}
%& Separation Fair & 0.63 & 6.83 & 6.16 &3.33 & 3.36 \\
%& Social Fair     & 0.68 & 0.48 & 0.44 & 0.02 & 0.02   \\
& \textsc{UniFair} & 0.66 & 1.16 & 1.12 &  0.06 &  0.07 \\
& DFC             & 0.69 & 248.08 & 243.08 &0.26  & 0.59 \\

%\midrule

%\multirow{4}{*}{MNIST-Reversed}
%& Separation Fair &  &  &  &  &  \\
%& Social Fair     &  &  &  &  &  \\
%& \textsc{UniFair} &  &  &  &  &  \\
%& DFC             &  &  &  &  &  \\

\bottomrule
\end{tabular}
\end{table}

As no prior deep fair clustering approach addresses our fairness notions, we compare with Deep Fair Clustering (DFC) \cite{li2020deep}. %Unlike our approach, 
DFC enforces independence between cluster assignments and the protected attribute, rather than explicitly optimizing the notions of separation fairness and social fairness. DFC learns fair representations through adversarial training, where an encoder produces clustering representations while a discriminator predicts the sensitive attribute. 
Table \ref{tab:deep_clustering_comparison} reports clustering quality (NMI), separation fairness, and social cost for our algorithm and the DFC method on the MNIST–USPS dataset, which is the only dataset with available pretrained models for this method.
For our algorithm we use the balanced setting $\lambda_{\text{soc}} = \lambda_{\text{sep}} = \lambda/2$. DFC achieves the highest clustering quality (NMI=$0.69$), though the performance of our methods remains very close. As the methods rely on different representation learning architectures, distance scales are not directly comparable, thus we focus on relative group differences.
%For the Separation fair approach, the separation fairness values for the two groups are very close,indicating similar distances from the decision boundaries.
%Similarly, the Social fair algorithm achieves identical social cost across the two groups, suggesting that they are equally distant from their cluster centers. Although each algorithm is optimizes a specific fairness objective, we also report the other metric for completeness. Notably, improving separation fairness does not significantly affect the social cost across groups.
The \textsc{UniFair} method shows balanced behavior, with minor differences among groups in both separation and social fairness values while preserving competitive clustering quality.
In contrast, DFC exhibits a noticeable discrepancy between the groups, particularly in the social cost metric, suggesting that although it achieves strong clustering performance and enforces independence between cluster assignments and the protected attribute, it does not satisfy the separation and social fairness notions considered in our work.

\section{Conclusions}
\label{sec:conclusions}

In this work, we proposed separation fair $k$-means, which encourages demographic groups to lie at similar distances from cluster decision boundaries, 
%and Social fair $k$-means, which reduces disparities in clustering cost across groups. 
%We further proposed 
and the \textsc{UniFair} algorithm, which jointly optimizes separation and social fairness. We
extended both approaches to deep clustering by enforcing these criteria in the latent space. 
Our experimental results on both tabular and high dimensional image data demonstrate that the proposed methods effectively improve separation fairness and reduce social cost while maintaining the clustering quality.
The results also show that separation fairness and social fairness capture different types of unfairness, and that the \textsc{UniFair} algorith allows flexible balancing between the two criteria. 
%In comparison with existing methods, we observe that no other approach explicitly focuses on these fairness criteria, while our method satisfies them while preserving clustering quality.
\bibliographystyle{splncs04}
\bibliography{bibliography}

@inproceedings{ghadiri2021socially,
  title={Socially fair k-means clustering},
  author={Ghadiri, M. and Samadi, S. and Vempala, S.},
  booktitle={ACM FAccT},
  pages={438--448},
  year={2021}
}

@inproceedings{abbasi2021fair,
  title={Fair clustering via equitable group representations},
  author={Abbasi, M. and Bhaskara, A. and Venkatasubramanian, S.},
  booktitle={ACM FAccT},
  pages={504--514},
  year={2021}
}

@inproceedings{makarychev2021approximation,
  title={Approximation algorithms for socially fair clustering},
  author={Makarychev, Y. and Vakilian, A.},
  booktitle={COLT},
  pages={3246--3264},
  year={2021}
}

@inproceedings{chierichetti2017fair,
  title={Fair clustering through fairlets},
  author={Chierichetti, F. and Kumar, R. and Lattanzi, S. and Vassilvitskii, S.},
  booktitle={NeurIPS},
  year={2017}
}

@article{ahmadi2020fair,
  title={Fair correlation clustering},
  author={Ahmadi, S. and Galhotra, S. and Saha, B. and Schwartz, R.},
  journal={arXiv},
  year={2020}
}

@article{schmidt2018fair,
  title={Fair coresets and streaming algorithms for fair k-means clustering},
  author={Schmidt, M. and Schwiegelshohn, C. and Sohler, C.},
  journal={arXiv},
  year={2018}
}

@inproceedings{bera2019fair,
  title={Fair algorithms for clustering},
  author={Bera, S. and Chakrabarty, D. and Flores, N. and Negahbani, M.},
  booktitle={NeurIPS},
  year={2019}
}

@article{anderson2020distributional,
  title={Distributional individual fairness in clustering},
  author={Anderson, N. and Bera, S. K. and Das, S. and Liu, Y.},
  journal={arXiv},
  year={2020}
}

@article{kleindessner2020notion,
  title={A notion of individual fairness for clustering},
  author={Kleindessner, M. and Awasthi, P. and Morgenstern, J.},
  journal={arXiv},
  year={2020}
}

@inproceedings{li2020deep,
  title={Deep fair clustering for visual learning},
  author={Li, P. and Zhao, H. and Liu, H.},
  booktitle={CVPR},
  pages={9070--9079},
  year={2020}
}

@article{wang2019towards,
  title={Towards fair deep clustering with multi-state protected variables},
  author={Wang, B. and Davidson, I.},
  journal={arXiv},
  year={2019}
}

@article{zhang2021deep,
  title={Deep fair discriminative clustering},
  author={Zhang, H. and Davidson, I.},
  journal={arXiv},
  year={2021}
}

@article{overview,
  title={An overview of fairness in clustering},
  author={Chhabra, A. and Masalkovaite, K. and Mohapatra, P.},
  journal={IEEE Access},
  volume={9},
  pages={130698--130720},
  year={2021}
}

@inproceedings{minirel,
  title={Fair minimum representation clustering},
  author={Lawless, C. and Gunluk, O.},
  booktitle={CPAIOR},
  series={LNCS},
  volume={14743},
  pages={20--37},
  year={2024}
}

@article{FPT,
  title={Tight FPT approximation for socially fair clustering},
  author={Goyal, D. and Jaiswal, R.},
  journal={Information Processing Letters},
  volume={182},
  pages={106383},
  year={2023}
}

@inproceedings{VardakasKPL25,
  title={Counterfactual explanations for k-means and Gaussian clustering},
  author={Vardakas, G. and Karra, A. and Pitoura, E. and Likas, A.},
  booktitle={ICTAI},
  pages={977--983},
  year={2025}
}
 
%% Note that this preceding line implies that you store your BibTeX references in a file called 'mybibliography.bib'. If you instead store your references in a file with a different name, for instance 'references.bib', the preceding line should read '\bibliography{references}'. Whatever you do, DO NOT put the file name extension .bib inside the \bibliography command; this will trip up LaTeX compilers. 
%
% If you do not want to use BibTeX, you can also type up the bibliography exactly as you see fit, using the following structure:
% Note that this number 8 reserves an amount of space (equal to the natural width of the given number) for the label of your references; if you have more than 9 references, you will want to change this number to 18. If you have more than 19 references, this number is best changed to 88. If you have more than 99 references, I salute you.
%Author, F.: Article title. Journal \textbf{2}(5), 99--110 (2016)

%Author, F., Author, S.: Title of a proceedings paper. In: Editor,
%F., Editor, S. (eds.) CONFERENCE 2016, LNCS, vol. 9999, pp. 1--13.
%Springer, Heidelberg (2016). \doi{10.10007/1234567890}

%Author, F., Author, S., Author, T.: Book title. 2nd edn. Publisher,

%Author, A.-B.: Contribution title. In: 9th International Proceedings
%on Proceedings, pp. 1--2. Publisher, Location (2010)

%LNCS Homepage, \url{http://www.springer.com/lncs}, last accessed 2023/10/25
%\end{thebibliography}
\end{document}